\documentclass[11pt,leqno]{article}
\usepackage{colacl}
\usepackage[english]{babel}
\usepackage{harvard}
\usepackage{fol}

\mathchardef\myand="2026

\title{Semantic interpretation of temporal information by abductive inference}
\author{\mbox{\shortstack{Sven Verdoolaege$^1$, Marc Denecker$^1$, Ness Schelkens$^2$, \\Danny De Schreye$^1$ and Frank Van Eynde$^2$}}\\
$^1$Department of Computer Science and $^2$Centre for
Computational Linguistics\\
K.U.Leuven}
\date{}

\begin{document}
\maketitle

\begin{abstract}
Besides temporal information explicitly available in verbs and
adjuncts, the temporal interpretation of a text also depends on general
world knowledge and default assumptions.  We will present a theory for
describing the relation between, on the one hand, verbs, their tenses and
adjuncts and, on the other, the eventualities and periods 
of time they represent
and their relative temporal locations.

The theory is formulated in logic 
and is a practical implementation of the concepts described
in Ness Schelkens et al.\ (this volume). We will show how an
abductive resolution procedure can be used on this representation to
extract temporal information from texts.
\end{abstract}

\section{Introduction}

This article presents some work conducted in the framework of
Linguaduct, a project on the temporal interpretation of Dutch texts by means
of abductive reasoning.

A natural language text contains a lot of both explicit and implicit
%Natural language texts contain a lot of both explicit and implicit
temporal information, mostly in verbs and adjuncts. 
The purpose of the theory presented here
is to represent how this information is available in Dutch texts
with the aim of allowing extraction of that information from particular
texts.% 
\footnote{In this paper, we only deal with sentences.}
The extracted information contains the temporal relations
between the eventualities described in the text as well as 
relations to periods of time explicitly or implicitly described
in the text. To arrive at this information some (limited)
reasoning needs to be performed on the representation.

The theory is an adaptation of the theory described in 
Van Eynde 1998\nocite{VanEynde1998}, Van Eynde 2000
\nocite{VanEynde2000} 
and Van Eynde (ms.)\nocite{VanEynde2000b},
which integrates a DRT-style analysis~\cite{Kamp1993} of tense and aspect
into HPSG~\cite{Pollard1994}.
Although from inception, HPSG was aimed at representing not only 
syntax, but also semantics and pragmatics, when it comes to
practical implementation, the focus has clearly been on syntax.
In contrast, the adapted theory deals mainly with semantics
and pragmatics. 
Furthermore, recent research suggests that a direct implementation
of HPSG may not be computationally viable \cite{Richter:Sailer:Penn:rsrl}.

For these reasons and because of our experience with it, we have
chosen to formulate the theory in logic, rather than trying to
extend an existing HPSG solution.
As such we do not use typed feature structure, but represent much
the same information in a different way.
A formulation in logic should also allow the linguistic knowledge
to be easily extended with additional world knowledge.
%A formulation in logic allows for an easy
%representation of both the linguistic knowledge and additional
%world knowledge and we can use an existing abductive procedure
%on the representation to perform the necessary reasoning.
The extraction of temporal information is performed by constructing
a model of the logical theory, for which we use an existing
abductive procedure. In case of ambiguities, several models exist
and each can be generated.

We start off with some preliminaries about the
adapted linguistic theory.
We will then describe the knowledge representation language
that we are going to use (essentially first order logic), 
followed by an extensive explanation
of the representation of the linguistic theory in logic.
We finish off with a short description of the reasoning procedure
used and some examples of how it can be used on the theory
to extract temporal information.

\section{Conceptual framework}

We assume familiarity with the following concepts
regarding the semantics of tense, aspect and temporal modification.
They correspond largely to what is found 
in Kamp and Reyle (1993)\nocite{Kamp1993}.

An {\em eventuality} is either an event, a state or a process.
We will not make use of these three kinds of eventualities,
but only of their defining properties, i.e.\ {\em stative}
vs.\ {\em dynamic} and {\em telic} vs.\ {\em atelic}.
The {\em eventuality time} is the period of time that 
the eventuality takes place on.
Eventualities are introduced by verbs, but not all verbs introduce
an eventuality. The ones that do are called {\em substantive}, those
that do not are called {\em vacuous} \cite{VanEynde2000}.

Each eventuality also introduces a {\em location time}.
The location time is a time with respect to which
the eventuality is located. It is similar to
Reichenbach's reference time \cite{Reichenbach1947} and
it is used here to express the difference in effect of
both tenses and frame adjuncts on 
the eventuality time of telic and atelic
eventualities.
It turns out that by mediating these effects through
the location time, only the relationship between
eventuality time and location time needs to be
telicity dependent.

The {\em utterance time} is the time within which the utterance
takes place and, in our theory, it is assumed to be constant
throughout the text.
To explain some phenomena (e.g.\ transposition%
%(See Kamp and Reyle (1993)
%chapter 5, section 4 or Van Eynde 1998, p.\ 245.)%
, flashback),
we use the {\em temporal perspective time},
which can change from clause to clause.%
\footnote{%We will not explain transpostion here. 
See Kamp and Reyle (1993)
chapter 5, section 4 or Van Eynde 1998, p.\ 245.}
We will from time to time refer to the temporal perspective time, but it
is not essential for understanding this paper.
In most circumstances, it can be assumed to be part of the
utterance time. It is only in the transposed case that it will
precede the utterance time.

As to adjuncts, our theory is mainly based on Schelkens et al.\ (this volume)%
\nocite{Schelkens2000}. We 
distinguish
 between
{\em frame} or {\em locating adjuncts} indicating
when an eventuality occurs, and {\em durational
adjuncts} expressing how long it takes.%
\footnote{We do not deal with frequency adjuncts in this paper.}
Amongst the frame adjuncts, we further 
distinguish
between {\em deictic} adjuncts that refer to the
utterance time (or more generally, to the temporal perspective time)
and {\em independent} adjuncts that refer directly
to the time axis.
In this paper, we do not deal with the so-called anaphoric
adjuncts.%, which refer to some previously mentioned period of time.

\section{Representation language}
\label{s:language}

As already mentioned, the theory is represented in
a language that is essentially first order logic (FOL).
It is extended with some axioms and notational conveniences,
which will be explained in this section.

First of all, we want different constants to represent
different entities and, more generally, we want different
functors and functors with differing arguments to represent
different entities. This means that constants and functors
identify objects; functors can be seen as constructors.
To accomplish this, the so-called
Unique Names Axioms (UNA) are added to the theory \cite{Reiter80a}
\cite{Clark78}.
Since there are an infinite number of such axioms,
they are (implicitly) built into the solver.

Sometimes, however, we want to use open functions,
that is, functions that do not identify an object, but
rather whose result can be equal to the result of another
function. 
The particular solver we use, assumes UNA for every functor,
so we cannot represent an $n$-ary
open function with a functor.
Instead we can use an $(n+1)$-ary predicate, with the extra
argument representing the result of the function and with
an axiom ensuring the existence and uniqueness of the function result
in function of its arguments.
For example, a binary function mapping a country and a year
to the person that is or was the president of that country
in that year, would be represented by a ternary predicate,
e.g.\ $\pres/(\USA/,2000,\Bill/)$.

We use a special notation for open functions that
also expresses that the arguments and the result
each satisfy a predicate. For example, the \pres/\ function
mapping a country and a year to a person is represented
as follows, where the {\tt{of}}\ indicates the introduction
of an open function.
\begin{equation}
\fol{of pres:: country(_), year(_) -> person(_).}
\end{equation}
The above declaration is equivalent to a set of axioms,
but it is shorter to write
down (and easier to understand at first glance) and it
allows reasoning procedures working on a
specification to handle such constraints more efficiently.
%The following is the more long-winded version:
The more long-winded version is shown in figure~\ref{f:pres}.
\begin{figure}
\begin{center}
%\begin{equation}
${\obeylines\fol{%
fol forall(C,Y)$ country(C) & year(Y) 
 => exists(P)$ person(P) & pres(C,Y,P).
fol forall(C,Y,P1,P2)$ pres(C,Y,P1) & 
 pres(C,Y,P2) => P1=P2.
fol forall(C,Y,P)$ pres(C,Y,P) 
 => country(C) & year(Y).
}}$
%\end{equation}
\caption{The open function \pres/.}
\label{f:pres}
\end{center}
\end{figure}
The {\tt fol} marker indicates that what follows is a first
order logic formula.
These axioms express,
first, that for each country and each year, there is a person that
is the president of that country in that year.
Second, that for a given country and year, there is at most one
president of that country in that year.
Third, that for each instantiation of the president predicate,
the first two arguments are a country and a year
respectively.

While FOL is ideally suited to represent assertional
knowledge, that is, facts and axioms, it does not fare
so well when it comes to definitional knowledge, i.e.\ to defining concepts.
A definition of a concept is an exhaustive enumeration of the
cases in which some object belongs to the concept.
We use a notation borrowed from logic programming.
For example, the concept {\it uncle} is defined as either 
a male sibling (i.e.\ brother) of a parent or a
male spouse of a (presumably female) sibling of a parent:
\begin{equation}
{\obeylines\fol{%
uncle(S,C) 
<- male(S) & sibling(S,P) & 
parent(P,C).

uncle(S,C) 
<- male(S) & married(S,A) & 
sibling(A,P) & parent(P,C).
}}
\end{equation}
Predicates that are not defined are called open predicates.
Open functions are always open predicates.

For the simple, non-recursive, definitions
we use in our theory, such a definition is equivalent to
its completion \cite{Clark78}.\footnote{The equivalence also
holds for a certain subset of recursive definitions.}
% (with $\leftarrow$ interpreted as an implication).
The completion of a definition states that 
a predicate defined in it,
holds if and only if one of its cases holds.
That is, the defined predicate, which is the head of
each rule, is equivalent to
the disjunction of the bodies of the rules.
In order to be able to take the disjunction of the bodies,
the heads have to be identical, of course.
To accomplish this, all terms
in an argument position of the predicate
in the head are first moved to the body as an equality to the variable
representing that argument, and all variables local to the
body are quantified in the body.
That is the set of
$m$ rules defining $p/n$: $\forall x_i:p(t_i)\leftarrow F_i$
is turned into the following equivalence:
\begin{equation}
\forall{\bf z}: p({\bf z}) \leftrightarrow \left \{ \begin{array}{c}
(\exists x_1: {\bf z=t}_1 \land F_1) \\
\lor\\
\ldots\\
\lor\\
(\exists x_m:\  {\bf z=t}_m \land F_m)
\end{array}\right.
\end{equation}
Denecker (2000)\nocite{Denecker2000c} presents the extension of
classical logic with a more general notion of (inductive)
definitions.

\section{Representation}

The way in which information can be extracted from a theory
is of course largely dependent on how the theory is represented.
We will therefore first discuss this representation and only
in the following sections will we present the extraction part.

\def\lb#1{$_{\hbox{\tiny #1}}$}

This section will mainly focus on the sentence ``[\lb{S} [\lb{NP} Ik] 
[\lb{VP} [\lb{V-AUX} ben] [\lb{VP} [\lb{ADV} gisteren]
[\lb{VP} [\lb{ADJ} ziek] [\lb{V-MAIN} geweest]]]]]'' 
(I have been sick yesterday),
showing how this sentence
is represented and showing the parts of the theory needed to extract
information from it.%
\footnote{Note that in general, there is no direct correspondence
between Dutch and English tenses.} 

\subsection{Input}
The sentence ``Ik ben gisteren ziek geweest'' contains three
interesting words when it comes to temporal information,
viz. ``ben'', ``geweest'' en ``gisteren''. The first
two are both forms of the verb ``zijn'' (to be), the last
is a temporal adjunct.

Since our main interest is in semantic interpretation,
we assume that any syntactic information that is
unambiguous and thus independent of any further 
semantic analysis, is available as input.
For this sentence, we have that ``geweest'' is the past
participle form of ``zijn'', and that it is the main verb
of the clause; ``ben'' is a present tense of ``zijn''
and is an auxiliary to ``geweest''; and finally,
``gisteren'' is an adjunct in the same clause.
Note that in general, when a clause contains more
than one (possibly) substantive verb, we do not
know a priori which of these verbs is modified by the adjunct.

To be able to refer to different occurrences of the
same word (such as ``zijn'' in the example), we make
use of tokens which are arbitrarily chosen constants
that represent these occurrences. In our example sentence
({\it s1}), we have verb tokens {\it w1} for the main
verb and {\it w2} for the auxiliary verb, and an adjunct
token {\it a1}. 

To express which word is associated with a token, we use
the \verbtword/\ and \adjtword/\ predicates. For verbs,
we additionally use the \morf/\ predicate to indicate the
verb form. Since all of these are exhaustive enumerations,
they are defined predicates.
For our example sentence we have:
\begin{equation}
{\obeylines\fol{%
verbt_word(w1,zijn) <- true.
verbt_word(w2,zijn) <- true.
adjt_word(a1,gisteren) <- true.
morf(w1,past_participle) <- true.
morf(w2,present_tense) <- true.
}}
\end{equation}

To express further that {\it s1} is a clause, 
that {\it w1} is the main verb of {\it s1},
that {\it w2} is an auxiliary verb with {\it w1} as its
complement and that {\it a1} is an adjunct in {\it s1},
we use the following:
\begin{equation}
{\obeylines\fol{%
clause(s1) <- true.
main_verb(s1,w1) <- true.
aux_verb(w2,w1) <- true.
s_adjunct(s1,a1) <- true.
}}
\end{equation}
Here again, we are dealing with exhaustive enumerations and thus
definitions.

\subsection{Eventualities and verbs}
\label{s:evt}

An eventuality associated with a verb token is indicated by the unary
\evt/\ functor. 
Not every such construct really represents
an eventuality, though. 
Only in the cases where the 
argument %is in fact %a verb token%, more particularly, when it
is a token for a substantive verb, does it refer to an
eventuality.
We therefore need an additional predicate \isevt/\ to indicate
which {\evt/}s represent eventualities.
\begin{equation}
{\obeylines\fol{%
isevt(evt(W)) <- subst(W).
}}
\end{equation}
%In the future, the theory might also deal with eventualities
%that are not directly tied to the occurrence of a verb,
%in which case this definition will have to be expanded.

Now we still have to define \subst/\ itself. There is only
a limited number of verbs that are not substantive, but we
cannot discriminate between substantive and vacuous verbs
on the basis of the lexeme. For example, ``zijn'' has (at least)
three uses: one as the main verb of a clause (\vzijn/, such as {\it w1}), 
one as a temporal auxiliary 
(\tzijn/, such as {\it w2}) and one as an aspectual auxiliary (\azijn/).
Of these, only the use as a temporal auxiliary is vacuous.
Section \ref{s:tense} deals with the distinction between
temporal and aspectual auxiliaries.

We consider these three ways of using ``zijn''
to be three different verb meanings of the same verb lexeme.
For each verb (meaning), the lexicon specifies 
whether it is substantive or not,
as well as any other property of verbs that is independent 
of the actual use of the verb in a sentence, e.g.\ whether
it is stative or dynamic. 
Since the properties of a verb hold for each occurrence of that
verb, they are inherited by the verb tokens.
%In general, we associate with
%verbs anything that pertains to the meaning.
%To be able to make the discrimination between
%substantive and vacuous, we introduce another
%type, \verb/, with which we associate not only substantivity 
%but also dynamicality. In general, anything that pertains
%to the meaning of a verb would be associated with entities of this type,
%but the two properties mentioned above are the only ones we use so far.
%The association between a verb token and the verb it is a token
%for is represented by the \tokenverb/\ predicate.

One of the properties of a verb, is the lexeme used.
The \verblex/\ predicate enumerates them and we naturally only
show part of its definition here.
\begin{equation}
{\obeylines\fol{%
verb_lex(t_zijn,zijn) <- true .
verb_lex(a_zijn,zijn) <- true .
verb_lex(v_zijn,zijn) <- true .
}}
\end{equation}
This property is not really inherited, but rather determines
of which verb a verb token can be an occurrence,
by placing a constraint on the open function \tokenverb/\ that
maps verb tokens to their corresponding verbs.
The constraint is that the verb is one of the possible
meanings of the word associated with the verb token,
i.e.\ that the word of the verb token is the same
as the word of the verb.
\begin{equation}
{\obeylines\fol{%
of token_verb:: verb_token(_) -> verb(_).
fol forall(T,V,W,L)$ token_verb(T,V) & 
verbt_word(T,W) & verb_lex(V,L).
 => W=L.
}}
\end{equation}
The \verbtoken/\ and \verb/\ predicates merely
enumerate the verb tokens and verbs respectively.
%and they are each defined in terms of a predicate known
%to already do the enumeration.
%These existing predicates are not unary
%predicates, though, and are therefore not usable 
%in an open function declaration.
%The predicate \verblex/\ defines which lexeme is associated with each verb
%and we naturally only show part of its definition here.
%%verb_token(T) <- exists(L)$ verbt_word(T,L).
%%verb(V) <- exists(L)$ verb_lex(V,L).

%With this \tokenverb/\ predicate, we can now define
%substantivity of a verb token in terms of substantivity
%of the associated verb. 
For each predicate representing one of the other properties
of verbs, there is a corresponding predicate for verb tokens,
that is defined to be that of the verb for which it is a token.
The predicates have the same name, with an extra {\it verb\_}
prefix for the one on verbs.
For example, substantivity of verb tokens (\subst/) is defined
as the substantivity of the corresponding verb (\verbsubst/).
\begin{equation}
{\obeylines\fol{%
subst(W) 
<- exists(V)$ token_verb(W,V) & 
verb_subst(V).
}}
\end{equation}
Due to the low number of vacuous verbs,
it is simpler to list those instead of all the substantive ones
and then to define the substantive verbs as those
that are not vacuous.
We only show part of the enumeration of vacuous verbs.
\begin{equation}
{\obeylines\fol{%
verb_subst(V) <- not verb_vacuous(V).
}}
\end{equation}
\begin{equation}
{\obeylines\fol{%
verb_vacuous(t_hebben) <- true.
verb_vacuous(t_zijn) <- true. 
verb_vacuous(t_zullen) <- true.
}}
\end{equation}

\subsection{Periods of time}

Part of the objective is to interrelate the different times
associated with a text, so it is necessary that they are all 
comparable to each other. 
The allowed kinds of periods of time are intervals, denoted
by the \int/\ predicate, and points, denoted by the \point/\ predicate.

Intervals refer directly to the time axis.
%Actually, we do not have a different representation
%for points (on the same level as the intervals,
%i.e.\ points that could be in, for instance, a before relation
%with an interval), they are just the smallest possible intervals.
%Having a separate representation would significantly
%complicate that part of the theory for little benefit.
They are represented by a pair of points on the time axis
as arguments to an \int/\ functor. This does not preclude
intervals from having no known end points. The points on the time
axis themselves can be left unspecified.
The relations between intervals (\overlap/, \within/, \before/\ and \meets/)
and some other properties about intervals that we will see later
(e.g.\ \daya/\ and \hour/) are defined in terms of relations and
properties of these end points. We will not show them here.

As to the actual representation of points on the time axis, it is not trivial
to find one that allows efficient processing for all the different
kinds of constraints that a text can place on the intervals that
it (implicitly or explicitly) deals with.
For now, we have contented ourselves with a fairly simple one
that is easy to process by humans.
Each point on the axis is represented by a function (\ts/) of
four integers, the year, the month, the day of the month
and the hour.\footnote{The rather low resolution is due to
technical limitations.} For example, the whole of May the 21st 1976
is represented as $\int/(\ts/(1976,5,21,0),\ts/(1976,5,22,0))$.
The relations and properties of these point are in turn
defined in terms of relations on their composing parts.

Most of the theory is not concerned with the exact representation
of periods of time, but instead only expresses relations between
pairs of such periods through the following predicates.
The relations that can be specified form only a subset of those
in Allen (1983)\nocite{Allen83}, but are sufficient for this application.
\begin{itemize}
\item\overlap/, which specifies that its
arguments have a non-empty intersection, 
\item\within/, a special
case of overlap which specifies that its first argument
is situated completely within its second argument,
\item\before/, which places its first argument completely
before its second argument, and
\item\meets/, a special case of before which specifies
that its first argument immediately precedes its second argument.
\end{itemize}

The eventuality time of an eventuality is associated with it
through the \evt/time\ predicate. Similarly, the location time
uses the \loc/\ predicate. The second argument of both is an
interval and is required to be unique for a given eventuality.
This is expressed by the following axioms:\footnote{The careful
reader will have noticed that these are prime candidates for
specification as
open functions. The current solver, however, places rather severe
restrictions on the domain predicate (\isevt/\ in this case,
as shown in the next set of axioms), 
which are not met here.}
\begin{equation}
{\obeylines\fol{%
fol forall(E,T)$ evttime(E,T) => int(T).
fol forall(E,L)$ loc(E,L) => int(L).
fol forall(E,T1,T2)$ evttime(E,T1) & 
 evttime(E,T2) => (T1=T2).
fol forall(E,L1,L2)$ loc(E,L1) & 
loc(E,L2) => (L1=L2).
}}
\end{equation}

Each eventuality has both an eventuality time and a location time
associated with it. Next, we express the overlap relation
that generally holds between these two. The second axiom expresses
the fact that for telic eventualities this overlap is narrowed down
to inclusion.
\begin{equation}
{\obeylines\fol{%
fol forall(E)$ isevt(E)  => (exists(L,T)$ 
loc(E,L)& evttime(E,T)&
overlap(T,L)).
fol forall(E)$ telic(E)  => (exists(L,T)$ 
loc(E,L)& evttime(E,T)&
within(T,L)).
}}
\end{equation}

The utterance time is expressed through the unary \utt/\ predicate
and is required to exist.
Essentially, it is just the eventuality time of some ``\utt/'' eventuality.
\begin{equation}
{\obeylines\fol{%
fol exists(U)$ utt(U).
utt(U) <- evttime(utt,U).
}}
\end{equation}

%Since the temporal perspective time can only change between clauses,
Since the temporal perspective time can only change from clause to clause,
each clause has a unique temporal perspective time and therefore
it can be represented as an (open) function of clauses.
\begin{equation}
{\obeylines\fol{%
of s_ppp:: clause(Z) -> point(_).
}}
\end{equation}
In the rest of the theory we will often want to refer
to the temporal perspective time of the clause that
contains a given verb. For convenience, we therefore
define another predicate \vppp/, relating verbs to 
temporal perspective times. 
In this definition
we use the \sverb/\ predicate, which enumerates all the verbs
in a clause.
%In this definition
%we use the \sverb/\ predicate, listing all the verbs
%in a clause, which are just the main verb and any (auxiliary) verb
%that has a verb in that clause as a complement.
\begin{equation}
{\obeylines\fol{%
v_ppp(W,P) 
 <- exists(Z)$ s_verb(Z,W) & 
   s_ppp(Z,P).
%
%s_verb(Z,W) 
% <- main_verb(Z,W).
%
%s_verb(Z,W) 
% <- exists(C)$ aux_verb(W,C) & 
%    s_verb(Z,C).
}}
\end{equation}

\subsection{Tenses and auxiliaries}
\label{s:tense}

The Dutch language has two simple tenses, the simple present and 
the simple past,
and several others that require auxiliaries.
Each is identified by a sequence of auxiliaries, each having
the next as a complement (the last has the main verb as
its complement), and the tense of the first auxiliary.
Actually, it is a sequence of classes of auxiliaries (e.g.\
an auxiliary of the perfect or an auxiliary of the future).
In our example sentence, ``Ik ben gisteren ziek geweest'',
there is one auxiliary ({\it w2}, ``zijn'') and it is one
of the perfect. The tense of this auxiliary is the present tense.

%What kind of auxiliary a verb functions as, is again
%information that we associate with a
%verb as defined in section~\ref{s:evt} (i.e.\ that to
%which meaning is assigned). The following shows
%part of the definition of the \verbauxkind/\ predicate.
Only some verbs are auxiliaries and there are several
kinds of auxiliaries (e.g.\ perfect, future).
This information is represented by the \verbauxkind/\ predicate.
For example, both uses of ``zijn'' as an auxiliary are auxiliaries of
the perfect.
\begin{equation}
{\obeylines\fol{%
verb_aux_kind(t_zijn,perfect) <- true .
verb_aux_kind(a_zijn,perfect) <- true .
}}
\end{equation}
As with the other properties of verbs, 
the auxiliary kind of a verb is inherited by
the verb tokens (as the \auxkind/\ predicate).
Here, we need to place an extra (indirect) constraint
on the \tokenverb/\ predicate.
%We then need to ensure that the verb tokens that operate
%as an auxiliary in the input are precisely those
%those that have this information associated with them.
%The following axiom accomplishes this.
The tokens that appear as an auxiliary in the sentence
(i.e.\ $\exists(\mbox{\it W2}) : \auxverb/(W,\mbox{\it W2})$)
should be precisely those whose verb is an auxiliary
(i.e.\ $\exists(F): \auxkind/(W,F)$).
\begin{equation}
{\obeylines\catcode`\$=12\fol{%
fol forall(W)$ (exists(W2)$ aux_verb(W,W2)) 
<=> (exists(F)$ aux_kind(W,F)).
}}
\end{equation}

This brings us to another distinction among auxiliaries 
made in the adapted theory
and that
is the one between temporal (e.g.\ \tzijn/) and aspectual 
(e.g.\ \azijn/) auxiliaries.
Temporal auxiliaries have a purely temporal meaning and
do not introduce an eventuality (i.e.\ they are vacuous).
Aspectual auxiliaries, on the other hand, have a temporal
influence only indirectly through the eventualities they
introduce. For example, the auxiliaries of the perfect
can be either temporal or aspectual.
The temporal perfect auxiliary requires the location time of the
verb it has as a complement to be situated before
the temporal perspective time, whereas the aspectual perfect
auxiliary introduces a new eventuality that represents
the resulting state of what its complement represents.

Both the temporal auxiliaries and the tenses stipulate a relation
between the location time of an eventuality and the
temporal perspective time. For temporal auxiliaries it is
the eventuality of the complement whose location time is constrained,
for the tenses it is the eventuality of the tensed verb,
in case it is substantive.

In our current theory, there are three kinds of relations
that can hold between a location time and the temporal perspective time,
which we have succinctly called \before/, \notbefore/\ and \after/.
Note that these names give only a rough idea. 
Further down in this section, we will show what exactly they imply.
It turns out that the temporal perfect and the past tense
have the same effect in this regard, namely the before relation.

There is a slight complication though, and that is that in case
of transposition, the past tense specifies a before relation
between the temporal perspective time and the utterance time,
between which, as you may remember, normally holds an inclusion relation.
Due to this, the transposed past tense does not specify a before
relation between location time and temporal perspective time, but
behaves rather like the present tense on this level, which we have
called the ``\notbefore/'' relation.
For completeness, we mention that the futurate auxiliary specifies
an after relation.
All this leads to the definition for the relation
between location time and temporal perspective time shown
in figure~\ref{f:evtppp}.
\begin{figure*}
\begin{center}
${\obeylines\fol{%
evt_ppp(W,before)<- subst(W) & morf(W,past_tense) & not transposed(W);
            (exists(A)$ aux_kind(A,perfect) & not subst(A) &  aux_verb(A,W)).
evt_ppp(W,not_before) <- subst(W) & (morf(W,past_tense) &transposed(W)); 
                 |phantom{subst(W) & (}morf(W,present_tense).
evt_ppp(W,after) <- exists(A)$ aux_kind(A,futurate) & aux_verb(A,W).
}}$
\caption{Relation between temporal perspective time and eventuality time.}
\label{f:evtppp}
\end{center}
\end{figure*}

As you can see, for each clause there is exactly one eventuality
whose location time is related to the temporal perspective time,
either that of the tensed verb in case it is substantive or that
of the complement of the (vacuous) temporal auxiliary.
You may wonder what happens when a temporal auxiliary has
the past tense. The above definition seems to allow that this
past tense has no effect (the present tense has indeed no effect,
but you would not be surprised about that). 
As it happens, the past tense for a temporal auxiliary, can only
be a transposed past tense and in this case the other relation,
i.e.\ the one between temporal perspective time and utterance
time comes into play, as we discussed above.
We give the definition of this relation and the axiom enforcing
the transposedness:
\begin{equation}
{\obeylines\fol{%
ppp_utt(W,before) 
<- morf(W,past_tense) &  transposed(W).

ppp_utt(W,at) 
<- morf(W,present_tense) ;
                 morf(W,past_tense) &               not transposed(W).
fol forall(W)$ morf(W,past_tense) & not subst(W) 
=> transposed(W).
}}
\end{equation}
We cannot go into the circumstances under which transposition
arises, we merely define it here as an ``abnormality'', where
\abnormal/\ is an open predicate:
\begin{equation}
{\obeylines\fol{%
transposed(W) 
<- abnormal(ab_transposed(W)).
}}
\label{e:transposed}
\end{equation}

As we already hinted at in section~\ref{s:evt}, the auxiliary ({\it w2})
in our example sentence can only be interpreted as a temporal auxiliary,
but we have shown no rule that would preclude it from being
interpreted as an aspectual one, simply because we have not shown
the rule for the aspectual perfect. So here it is:
\begin{equation}
{\obeylines\fol{%
fol forall(W,C)$ aux_kind(W,perfect) & 
 subst(W) & aux_verb(W,C)
        => not stative(evt(C)) & 
	result(evt(C),evt(W)).
}}
\end{equation}
As already explained, the axiom states that the aspectual (substantive)
perfect introduces the resulting state of the eventuality of its
complement. Additionally, in the theory, there is a requirement
for the complement to be non-stative, which is why {\it w2} cannot
be aspectual, since in our example
sentence, {\it w1} is stative.
The definition of the \stative/\ predicate is similar to the \subst/\ one
shown in section~\ref{s:evt}.

The \result/\ predicate in the above axiom is an open predicate
expressing that the second argument represents the resulting
state of the first argument. This implies that the resulting
state immediately follows the eventuality that it is a resulting
state for.
\begin{equation}
{\obeylines\fol{%
fol forall(E,E2,L,L2)$ result(E,E2) & 
 evttime(E,L) & evttime(E2,L2)
        => meets(L,L2).
}}
\end{equation}

Now, as promised, the implications of the \evtppp/\ and \ppputt/\
relations. For the \ppputt/\ relation, \before/\ means exactly
what you would expect and \at/\ means inclusion of the
temporal perspective time in the utterance time.
\begin{equation}
{\obeylines\fol{%
fol forall(U,W,P)$ ppp_utt(W,before) & 
 v_ppp(W,P) & utt(U)
        => before(P,U).
fol forall(U,W,P)$ ppp_utt(W,at) & 
 v_ppp(W,P) & utt(U)
        => within(P,U).
}}
\end{equation}
For the \evtppp/\ relation, \notbefore/\ means non precedence.
(There are some further specialisations of this relation
in restricted circumstance, but we will not show them here.)
The \before/\ and \after/\ cases should be intuitively clear.
\begin{equation}
{\obeylines\fol{%
fol forall(W,P,L)$ evt_ppp(W,before) & 
        loc(evt(W),L) & v_ppp(W,P)
        => before(L,P).          
fol forall(W,P,L)$ evt_ppp(W,not_before) &
        loc(evt(W),L) & v_ppp(W,P)
        => not before(L,P).      
fol forall(W,P,L)$ evt_ppp(W,after) & 
         loc(evt(W),L) & v_ppp(W,P)  
        => before(P,L).
}}
\end{equation}

\subsection{Adjuncts}

Although we currently deal with only a few adjuncts,
we have tried to be as general as possible in our
treatment of them. 
%This has resulted in a predicate
%\timeperiod/\ that may seem rather complicated at first
%sight, but we hope to show here that it may have some merits.
Our main design decisions were to have compositionality
and to avoid redundancy.
This means that we associate with an adjunct (token)
only what the adjunct itself adds to the meaning of the
sentence and we want to treat what is common to several
adjuncts, separately.

For example, the adjunct ``gisteren'' (yesterday),
is a frame adjunct that refers to the day before
the day that includes the temporal perspective time.
That is, the adjunct specifies a {\em relation} between
some external period of time (in case of deictic
adjuncts such as ``gisteren'', the temporal perspective time) and
the period of time that it introduces.
Anaphoric adjuncts, which, as we already mentioned, we do not deal
with in this paper, specify a relation between some other period
of time and the period of time that they introduce.

We therefore generalise that for frame adjuncts, the meaning
is this relation between a 
contextual parameter and the introduced period of time.
To represent this, a tuple consisting of the two periods of time
is associated with the adjunct and the relation between them
is enforced by an axiom. This is the axiom that expresses
the ``gisteren'' relation as explained above:
\begin{equation}
{\obeylines\fol{%
fol forall(A)$ adjt_word(A,gisteren)
        => (exists(P,Y,T)$ day_a(Y)& day_a(T)& 
 within(P,T) & meets(Y,T) & 
        time_period(A,ta_frame(deictic(P),Y))).
}}
\end{equation}
The \taframe/\ functor specifies that the adjunct is a frame adjunct.
Its first argument, the contextual parameter, is a functor \deictic/,
which indicates that its only argument (ultimately) refers to the
temporal perspective time of some clause. Note that the axiom
only requires the existence of this period $P$, which only
at a later stage will be identified with some temporal perspective time,
by an axiom shown later in this subsection.
Independent frame adjuncts will have
the constant \independent/\ instead of the \deictic/\ functor.

In general, each adjunct has some (unique) construct associated with it.
This means that \timeperiod/\ is an open function of an adjunct.
Since the different axioms specify what the function result is,
it is left unspecified in the declaration of \timeperiod/\ as an open function.
\begin{equation}
{\obeylines\fol{%
adjunctoid(A) <- exists(L)$ adjt_word(A,L).
anything(_) <- true.
of time_period:: adjunctoid(_) -> anything(_).
}}
\end{equation}
As you can see, we do not only associate a time period
with adjuncts, but with anything that ``looks'' like
an adjunct (\adjunctoid/), 
which is anything that occurs as a first argument to
the \adjtword/\ predicate.
We will show the use of that later in this subsection.

The effect of a frame adjunct is that the location time of the
modified verb is required to be within the frame period:
\begin{equation}
{\obeylines\fol{%
fol forall(W,T,R,L)$
    verb_time(W,ta_frame(R,T)) & loc(evt(W),L) 
	=>  within(L,T).
}}
\end{equation}
Here, again, we use an auxiliary predicate \verbtime/, which
associates the time period information to the verb that is
modified:
\begin{equation}
{\obeylines\fol{%
verb_time(W,P) 
<- exists(A)$ verb_adjunct(W,A) & 
time_period(A,P).
}}
\end{equation}
When it is a deictic adjunct that modifies a verb, we still have to
require that the contextual parameter is in fact the
temporal perspective time:
\begin{equation}
{\obeylines\fol{%
fol forall(V,A,K,P)$ 
 verb_time(V,ta_frame(deictic(A),K)) & 
 v_ppp(V,P) => A = P.
}}
\end{equation}
 
The \timeperiod/\ also allows for some more complicated uses,
for example, {\em na} (after) followed by something that
could be used as a frame adjunct. The use of {\em na} 
indicates that the location time of the modified verb
is situated after the period of time of the complement
of the preposition.
This can 
be modelled as inclusion in a frame interval that immediately
succeeds the frame interval of the complement. 
The contextual dependency is determined by the complement
and therefore it is left unchanged.
\begin{equation}
{\obeylines\fol{%
fol forall(A,B,C,F)$ adjunct(A,na(B)) & 
 time_period(B,ta_frame(C,F))
=> (exists(T)$ meets(F,T) & 
time_period(A,ta_frame(C,T))).
}}
\end{equation}
In the input, {\em na gisteren} would be specified as follows:
\begin{equation}
{\obeylines\fol{%
adjt_word(a1,gisteren) <- true.
adjt_word(a2,na(a1)) <- true.
s_adjunct(s1,a2) <- true.
}}
\end{equation}

As for durational adjuncts, we can only represent so called in-adjuncts
so far, and not the for-adjuncts. The cause of this is that for-adjuncts
require the eventuality of the verb they modify to be atelic in itself,
but combined with the for-adjunct, the eventuality would be telic.
Now, one and the same eventuality can obviously not be both
telic and atelic, so the adjunct would have to introduce
a new eventuality and we do not handle such things yet.

Handling {\em in} is done in much the same way as {\em na},
except that there is a change in type instead of a
change in interval, since the complement of
{\em in} when used by itself is not an in-adjunct.
\begin{equation}
{\obeylines\fol{%
fol forall(A,B,C,T)$ adjunct(A,in(B)) & 
 time_period(B,ta_dur(T))
=> time_period(A,ta_in(T)).
}}
\end{equation}
An in-adjunct requires its corresponding eventuality to have taken
at most as much time as specified by its complement.
\begin{equation}
{\obeylines\fol{%
fol forall(W,T,L)$ 
    verb_time(W,ta_in(T)) & evttime(evt(W),L) 
 =>  telic(evt(W)) & within(L,T).
}}
\end{equation}
This is again similar to the frame adjunct constraint, the difference
being that it requires a telic eventuality and 
that, here, the eventuality time is constrained directly,
instead of indirectly through the location time.
The location time {\em could} of course be used here as well,
since, for telic eventualities, the eventuality time is restricted to be within the location
time, but there is no need to here.

An example of a complement of {\em in} would be {\em een uur} (an hour):
%which we show here only schematically.
\begin{equation}
{\obeylines\fol{%
fol forall(A)$ adjt_word(A,een_uur)
        => (exists(T)$ hour(T) &
        time_period(A,independent,ta_dur(T))).
}}
\end{equation}
Here, the \hour/\ predicate means that its argument has a duration
of exactly one hour.

\section{Reasoning procedure}
\label{s:procedure}

Up to this point, we have presented a logic theory consisting
of FOL axioms and definitions in which some predicates were
defined and others open, but we have not shown how to derive
information from this theory. Since the defined predicates
are known, the only uncertainty lies in the open predicates.
We need to find an interpretation for these open predicates
that is consistent with the theory. In other words,
we need to generate a model for the open predicates.
To do this we use an existing abductive procedure,
called SLDNFA \cite{VanNuffelen99}, which operates on theories written in
the knowledge representation language outlined in 
section~\ref{s:language}.

%\def\verschil#1#2{a\newcount\d\d=#2\advance\d by -#2\the\d b}
%\newcount\d
%\def\verschil#1#2{a\d=2\number\d b}
%\def\quux#1#2{a
%\d=#1
%\advance\d by -#2
%\number\d
%b
%}
%\quux{\thepage}{1}
%\tracingmacros=1
%\edef\bar{\pageref{e:transposed}}
%\tracingmacros=0

Abduction is a form of non-monotonic reasoning that
is used to explain observations. In this case,
an explanation consists of a model for the open predicates.
As a trivial example, take the definition of \transposed/\
in section~\ref{e:transposed}. Suppose we somehow observe
$\transposed/(\evt/({\it w1}))$, then we can explain this
by assuming (abducing) $\abnormal/(\abtransposed/(\evt/({\it w1})))$.
This form of reasoning is called non-monotonic, because new
information may invalidate previously drawn conclusions.
(This in contrast to deduction, where any conclusion that was
drawn, can always still be drawn after the addition of new information.)
For example, should we add the axiom
\begin{equation}
{\obeylines\fol{%
fol not abnormal(ab_transposed(evt(w1))).
}}
\end{equation}
then we would not be able to draw the same conclusion,
as it would be inconsistent.
(This is not a problem for deduction, since we can deduce
anything from an inconsistent theory.)

We cannot give a detailed explanation of how the SLDNFA procedure
works within this limited space, but we do want to give an idea of 
it.
In short, the procedure tries to make the conjunction of
all axioms and the (possibly empty) observation (query) hold
by abducing a set of atoms, according to the following rules. 
A conjunction holds if all of its
conjuncts hold; for a disjunction, it is sufficient that one of
its disjuncts holds, so each one is tried out until one is found
that holds. Negation is distributed over disjunctions and conjunctions. 
A defined predicate is replaced by its completion 
(see section~\ref{s:language}).

When an open predicate occurs negatively, the atom is 
(temporarily) assumed not
to hold if this does not conflict with earlier made assumptions.
If it is part of a disjunction, the remaining disjuncts
are remembered in case it turns out we want it to hold after all.
When an open predicate occurs positively, the atom is assumed to
hold and any applicable remembered disjunction is required
to hold as well. If the atom was already assumed not to hold
and there were no remaining disjuncts at that point,
the procedure backtracks to the latest disjunction with remaining
disjuncts. Finally, an equality unifies its arguments. If this fails,
the procedure backtracks as well.

Numerical operations and comparisons are
translated into CLP (Constraint Logic
Programming) constraints and handed over to
an efficient constraint solver.
They occur mainly in the definitions and axioms
pertaining to points on the time axis, which we
have not shown in this paper.
At the end, all numerical variables are
labelled, which means that they
get a value assigned to them that satisfies
all constraints.

\section{Deriving temporal information}

Applying the procedure of section~\ref{s:procedure} to our
theory with an empty observation (that is, we just want a
consistent interpretation for the open predicates)
for our example sentence ``Ik ben gisteren ziek geweest'',
we get the following result, which lists for each open predicate
precisely its model:

\begin{verbatim}
adjunct_verb : [adjunct_verb(a1,w1)]
evttime : 
[evttime(evt(w1),int(ts(1999,1,1,0),
                     ts(1999,1,2,0))),
 evttime(utt,int(ts(1999,1,1,0),
                 ts(1999,1,3,0)))]
loc : 
[loc(evt(w1),int(ts(1999,1,1,0),
                 ts(1999,1,2,0)))]
s_ppp : [s_ppp(s1,int(ts(1999,1,2,0),
                      ts(1999,1,2,1)))]
time_period : [time_period(a1,
 ta_frame(deictic(
  int(ts(1999,1,2,0),ts(1999,1,2,1))),
  int(ts(1999,1,1,0),ts(1999,1,2,0))))]
token_verb : [token_verb(w2,t_zijn),
              token_verb(w1,v_zijn)]
\end{verbatim}

The {\it w1} and {\it w2} tokens were abduced to be tokens for
the verb \vzijn/\ and the temporal auxiliary \tzijn/, respectively.
As explained in section~\ref{s:tense} this is in fact the only
possibility for the \tokenverb/\ predicate for this particular
sentence. The only adjunct ({\it a1}) in the sentence was taken to 
modify the main verb ({\it w1}), which is again the only
possibility here, since the temporal auxiliary is vacuous.
This model is unique upto the choice of the intervals.
In case of ambiuguities, several models are generated.

Since no information was given about the time, the time intervals are
chosen rather arbitrarily, although they do of course satisfy
the constraints. The temporal perspective time is the first
hour of January the second, 1999. It is included in the utterance
time, which,
as you can see, is rather large, stretching over two days.%
%it does of course conform to the theory.%
\footnote{Maybe some general restriction
should be placed on the extent of the utterance time.}
The temporal perspective time is further the deictic argument
relative to which ``gisteren'' (yesterday) is positioned.
This January the first includes the location time of
the eventuality associated with {\it w1} (the whole day, here),
which in turn includes the eventuality time (also the whole day, here).

Suppose that you know that the sickness lasted from 6 o'clock until
8 o'clock in the evening on May the 21st, 2000 and that the
utterance time lasted an hour, you would give the following
observation (query) to the system:
\begin{equation}
{\obeylines\fol{%
utt(U) &hour(U) &
 evttime(evt(w1),int(ts(2000,5,21,18)|phantom{))}
 |phantom{evttime(evt(w1),int(}ts(2000,5,21,20)))
}}
\end{equation}
The result is, as you would expect, that the utterance happened on
May the 22nd. We only show the result for the \evt/time\ predicate;
the other time intervals are changed accordingly.
\begin{verbatim}
evttime : 
[evttime(utt,int(ts(2000,5,22,0),
                 ts(2000,5,22,1))),
evttime(evt(w1),int(ts(2000,5,21,18),
                    ts(2000,5,21,20)))]
\end{verbatim}

An example of another kind of query, is the following.
Is it possible for the utterance to have taken place
before the eventuality described by {\it w1} ?
\begin{equation}
{\obeylines\fol{%
utt(U) & evttime(evt(w1),T) & 
before(U,T)
}}
\end{equation}
The result:
\begin{verbatim}
no
\end{verbatim}

\section{Conclusions}

In this paper, we have shown a formalisation 
in first order logic
of an existing
theory about temporal information in Dutch texts.
Although to a large extent, this theory had already been formalised
in HPSG, it had not been practically implemented as such.

The representation shown in this paper has effectively been
implemented and results of  using an abductive reasoning 
procedure on it were presented.
Although the problems dealt with in this paper were limited,
the experiments show that abduction may be viable for
natural language processing.

A representation in logic is not only very flexible
and extensible, a good representation also requires the use of well
thought out concepts, because of logic's clear
and formal semantics.
This representation has already helped in clearing out
the meaning of some concepts used in the temporal analysis
of sentences.

The research that led to this paper has also been
an exercise in knowledge representation and will
contribute towards a better knowledge representation
methodology.

\bibliography{clin99b}
\bibliographystyle{dcu}

\end{document}